# Union is strength in lossy image compression

Mario Mastriani

*Abstract*—In this work, we present a comparison between different techniques of image compression. First, the image is divided in blocks which are organized according to a certain scan. Later, several compression techniques are applied, combined or alone. Such techniques are: wavelets (Haar's basis), Karhunen-Loève Transform, etc. Simulations show that the combined versions are the best, with minor Mean Squared Error (MSE), and higher Peak Signal to Noise Ratio (PSNR) and better image quality, even in the presence of noise.

## I. Introduction

MODERN image compression techniques often involve Discrete Wavelet Transform (DWT) [1-18] with different scans for the wavelets subbands [19-26] and Karhunen-Loève Transform (KLT) [27-29]. While DWT is applied to image compression [8-11,13-16], KLT is applied in image decorrelation [30-34], that is to say, KLT is used inside compression techniques of several images with a high degree of mutual correlation, for example, frames of medical images [35], video [36, 37], and multi [30, 32-34] and hyperspectral imagery [38-40].

Many efforts have been made in the recent years in order to compress efficiently such data sets. The challenge is to have a data representation which takes into account at the same time both the advantages and disadvantages of KLT [29], for a most efficient compression based on an optimal decorrelation.

Several authors have tried to combine the DWT with the KLT but with questionable success [1], with particular interest to multispectral imagery [30, 32, 34].

In all cases, the KLT is used to decorrelate in the spectral domain. All images are first decomposed into blocks, and each block uses its own KLT instead of one single matrix for the whole image. In this paper, we use the KLT for a decorrelation between sub-blocks resulting of the applications of a DWT with different scans, that is to say, in the wavelet domain.

We introduce in this paper an appropriate sequence, decorrelating first the data in the spatial domain using the DWT (Haar's basis) and afterwards in wavelet domain, using the KLT, allows us a more efficient (and robust, in presence of noise) compression scheme.

The resulting compression scheme is a lossy image compression. This type of compression system does not retain the exact image pixel to pixel. Instead it takes advantage of limitations in the human eye to approximate the image so that it is visually the same as the original. These methods can achieve vastly superior compression rates than lossless methods, but they must be used sensibly [41].

Lossy compression techniques generally only work well with real-life photography; they often give disastrous results with other types of images such as line art, or text. Putting an image through several compression-decompression cycles will cause the image to deteriorate beyond acceptable standards. So a lossy compression should only be used after all processing has been done, it should not be used as an intermediate storage format. Further note that while the reconstructed image looks the same as the original, this is according to the human eye. If a computer has to process the image in a recognition system, it may be completely thrown off by the changes [41].

On the other hands, consider the generic transform coder in Fig.1 consisting of a 2-D transform, quantizer, and entropy coder. We see here that loss occurs during quantization and after the transform. Therefore, in order to conduct our analysis, we must repeat the transform to return to the stage where loss occurs and examine the effect of quantization on transform coefficients [42].

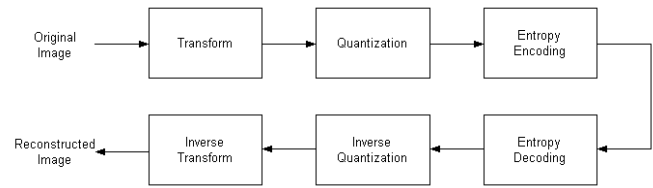

Fig. 1 Generic transform coding for digital images

In this work, additional losses are incorporated, because, after of KLT applications a pruning of decorrelated sub-blocks is applied before the quantization, with a statistical criterion [28].

The Bidimensional Discrete Wavelet Transform and the method to reduce noise and to compress by wavelet thresholding is outlined in Section II. Scans are outlined in Section III. KLT is outlined in Section IV. Combinations are outline in Section V. In Section VI, we discuss briefly the more appropriate metrics for compression. In Section VII, the experimental results using the proposed algorithm are presented. Finally, Section VIII provides a conclusion of the paper.

## II. BIDIMENSIONAL DISCRETE WAVELET TRANSFORM

The Bidimensional Discrete Wavelet Transform (DWT-2D) [6], [7], [8]-[16], [43]-[56] corresponds to multiresolution approximation expressions. In practice, mutiresolution analysis is carried out using 4 channel filter banks (for each level of decomposition) composed of a low-pass and a high-pass filter and each filter bank is then sampled at a half rate (1/2 down sampling) of the previous frequency. By repeating this procedure, it is possible to obtain wavelet transform of any order. The down sampling procedure keeps the scaling parameter constant (equal to ½) throughout successive wavelet transforms so that is benefits for simple computer implementation. In the case of an image, the filtering is implemented in a separable way be filtering the lines and columns.

Note that [6], [7] the DWT of an image consists of four frequency channels for each level of decomposition. For example, for $i$-level of decomposition we have:

$LL_{n,i}$: Noisy Coefficients of Approximation.
$LH_{n,i}$: Noisy Coefficients of Vertical Detail,
$HL_{n,i}$: Noisy Coefficients of Horizontal Detail, and
$HH_{n,i}$: Noisy Coefficients of Diagonal Detail.

The LL part at each scale is decomposed recursively, as illustrated in Fig. 2 [6], [7].

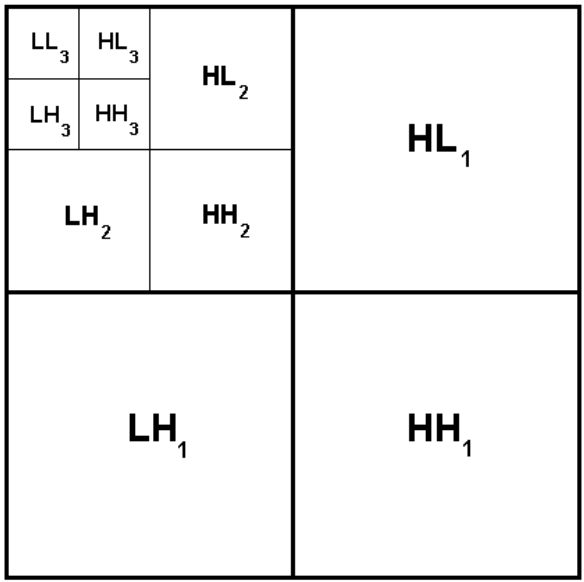

Fig. 2 Data preparation of the image. Recursive decomposition of LL parts.

To achieve space-scale adaptive noise reduction, we need to prepare the 1-D coefficient data stream which contains the space-scale information of 2-D images. This is somewhat similar to the "zigzag" arrangement of the DCT (Discrete Cosine Transform) coefficients in image coding applications [46]. In this data preparation step, the DWT-2D coefficients are rearranged as a 1-D coefficient series in spatial order so that the adjacent samples represent the same local areas in the original image [48].

Fig.3 shows the interior of the DWT-2D with the four subbands of the transformed image [55], which will be used in Fig.4. Each output of Fig. 3 represents a subband of splitting process of the 2-D coefficient matrix corresponding to Fig. 2.

### A. Wavelet Noise Thresholding

The wavelet coefficients calculated by a wavelet transform represent change in the image at a particular resolution. By looking at the image in various resolutions it should be possible to filter out noise, at least in theory. However, the definition of noise is a difficult one.

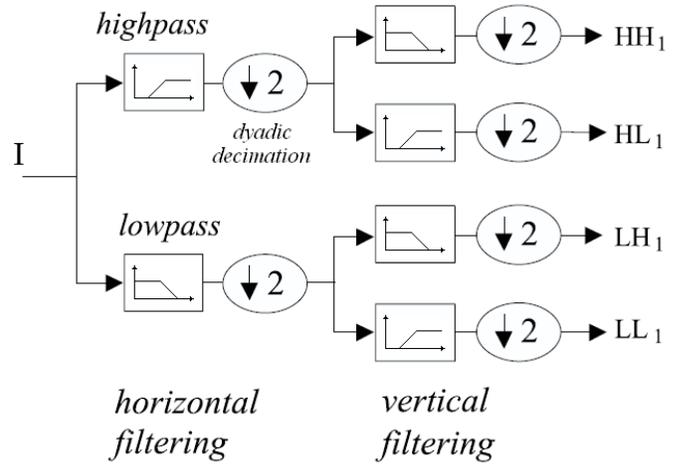

Fig. 3 Two dimensional DWT. A decomposition step. Usual splitting of the subbands.

In fact, "one person's noise is another's signal". In part this depends on the resolution one is looking at. One algorithm to remove Gaussian white noise is summarized by D.L. Donoho and I.M. Johnstone [2], [3], and synthesized in Fig. 4.

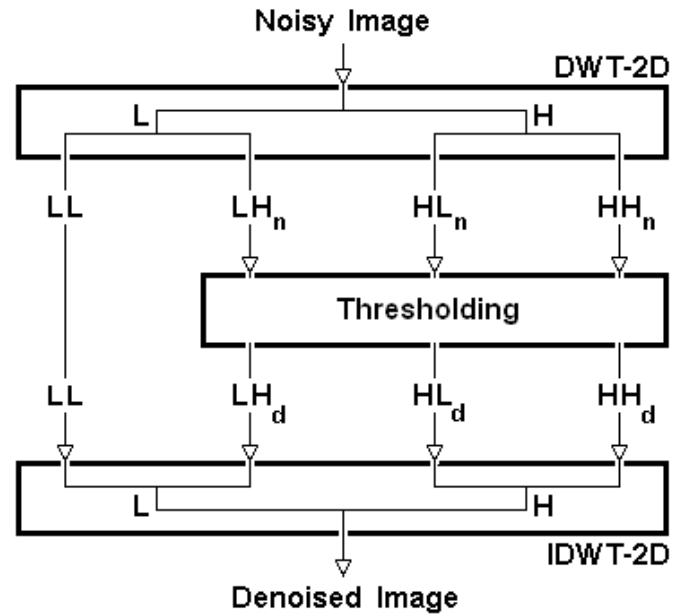

Fig. 4 Thresholding Techniques

The algorithm is:
1) Calculate a wavelet transform and order the coefficients by increasing frequency. This will result in an array containing the image average plus a set of coefficients of length 1, 2, 4, 8, etc. The noise threshold will be calculated on the highest frequency coefficient spectrum (this is the largest spectrum).
2) Calculate the *median absolute deviation* (mad) on the largest coefficient spectrum. The median is calculated from the absolute value of the coefficients. The equation for the median absolute deviation is shown below:

$$\delta_{mad} = \frac{median(|C_{n,i}|)}{0.6745} \quad (1)$$

where $C_{n,i}$ may be $LH_{n,i}$, $HL_{n,i}$, or $HH_{n,i}$ for *i*-level of decomposition. The factor 0.6745 in the denominator rescales the numerator so that $\delta_{mad}$ is also a suitable estimator for the standard deviation for Gaussian white noise [5], [46], [47].

3) For calculating the noise threshold $\lambda$ we have used a modified version of the equation that has been discussed in papers by D.L. Donoho and I.M. Johnstone. The equation is:

$$\lambda = \delta_{mad}\sqrt{2\log[N]} \quad (2)$$

where N is the number of pixels in the subimage, i.e., HL, LH or HH.

4) Apply a thresholding algorithm to the coefficients. There are two popular versions:

4.1. Hard thresholding. Hard thresholding sets any coefficient less than or equal to the threshold to zero, see Fig. 5(a).

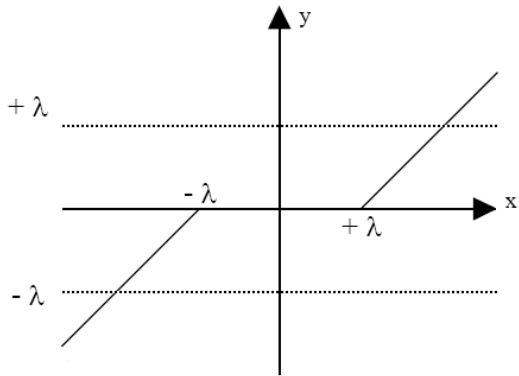

Fig. 5(a) Soft-Thresholfing

where *x* may be $LH_{n,i}$, $HL_{n,i}$, or $HH_{n,i}$, *y* may be $HH_{d,i}$ : Denoised Coefficients of Diagonal Detail,
$HL_{d,i}$ : Denoised Coefficients of Horizontal Detail,
$LH_{d,i}$ : Denoised Coefficients of Vertical Detail,
for *i*-level of decomposition.

The respective code is:

```
for row = 1:N
  for column = 1:N
    if |C_n,i[row][column]| <= λ,
      C_n,i[row][column] = 0.0;
    end
  end
end
```

4.2. Soft thresholding. Soft thresholding sets any coefficient less than or equal to the threshold to zero, see Fig. 5(b). The threshold is subtracted from any coefficient that is greater than the threshold. This moves the image coefficients toward zero.

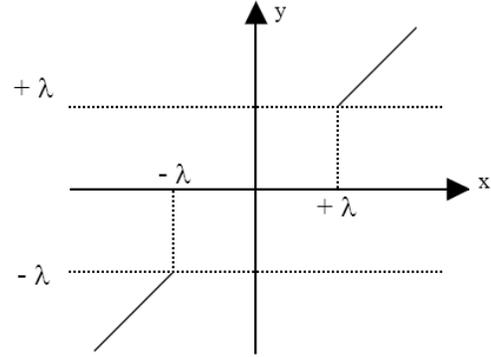

Fig. 5(b): Hard-Thresholfing

The respective code is:

```
for row = 1:N
  for column = 1:N
    if |C_n,i[row][column]| <= λ,
      C_n,i[row][column] = 0.0;
    else
      C_n,i[row][column] = C_n,i[row][column] - λ;
    end
  end
end
```

III. SCANS

Fig.6 shows four different types of spatial scanning methods [57, 58]. In this paper we use (a) and (d) scans.

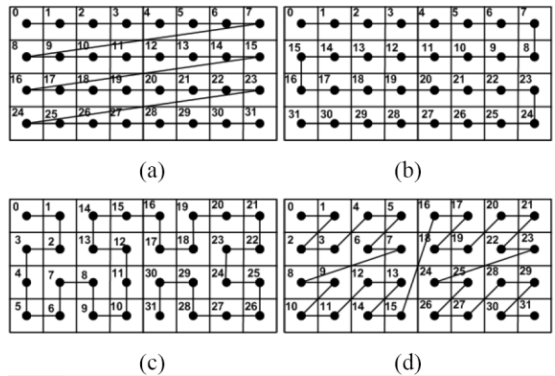

(a)      (b)

(c)      (d)

Fig. 6: Different space scanning methods. a) Row (Raster) order, b) Row prime order, c) Peano-Hilbert order, d) Morton (Z) order

In Fig.6 each numbering cell represent a sub-block (inside wavelet domain) which may be spatially ordered (in upward order) in a three dimensional matrix before KLT, see Fig.7.

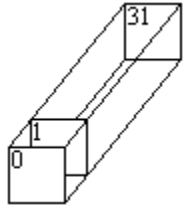

Fig. 7: Building of 3D-matrix with sub-blocks in upward order

As can be seen from Fig.6, pixels, which have to be treated or not with a DWT, are concentrated in blocks. Block clusters of 2×2, 4×4, 8×8 … pixels, can be easily extracted, since pixels in these blocks are transmitted one after another (row ordering does not posses this valuable feature because pixels are transmitted serially row after row). This feature can be handy for spatial image processing, such as resolution reduction. In order to reduce image resolution by a factor of two, the mean of four pixels (a 2×2 block) has to be calculated. With these orderings (Morton and Row-rafter), it can be done in a simple, straightforward way, without requiring multiple storage elements. This calculation can be expanded to blocks of sizes 4×4, 8×8 etc.

## IV. KARHUNEN-LOEVE TRANSFORM (KLT)

The KLT begin with the covariance matrix of the vectors **x** generated between values of pixel with similar allocation in all arranged sub-blocks of 3D-matrix, as show in Fig.8.

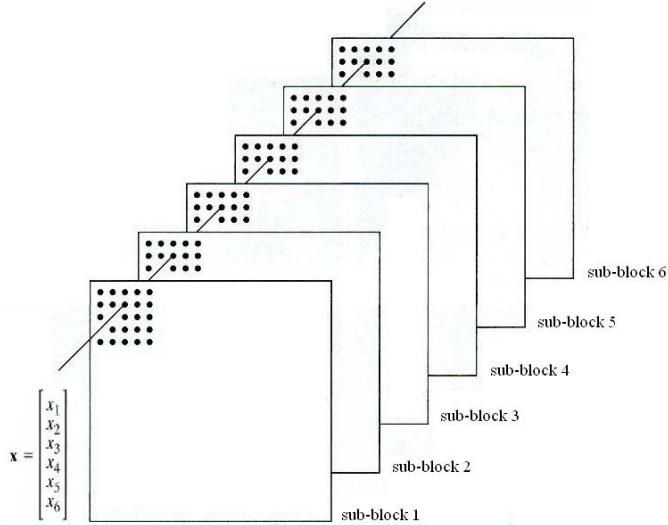

Fig. 8: Formation of a vector from corresponding pixels in six sub-blocks

The covariance matrix results,

$$C_x = E\{(x-m_x)(x-m_x)^T\} \quad (3)$$

with:

$\mathbf{x} = (x_1, x_2, \ldots, x_n)^T$, where **x** is one of the correlated original vector set, "T" indicates transpose and $n$ is the number of sub-blocks.

$\mathbf{m_x} = E\{\mathbf{x}\}$ is the mean vector, and where $E\{\bullet\}$ is the expected value of the argument, and the subscript denotes that **m** is associated with the population of **x** vectors.

In the appropriate mathematical form:

$$m_x = \frac{1}{rsb*csb} \sum_{k=1}^{rsb*csb} x_k \quad (4)$$

where:
rsb is the sub-block row number
csb is the sub-block column number

On the other hands,

$$C_x = \frac{1}{rsb*csb} \sum_{k=1}^{rsb*csb} (x_k - m_x)(x_k - m_x)^T \quad (5)$$

Therefore, KLT will be,

$$\mathbf{y} = \mathbf{V}^T (\mathbf{x}\text{-}\mathbf{m_x}) \quad (6)$$

with:
$\mathbf{y} = (y_1, y_2, \ldots, y_n)^T$, where **y** is one of the decorrelated transformed vector set
**V** is a matrix whose columns are the eigenvectors of $\mathbf{C_x}$.

When applying the calculus of eigenvectors, two matrices arise, **V** y $\mathbf{C_y}$, being $\mathbf{C_y}$ a diagonal matrix, where the elements on its main diagonal are de eigenvalues of $\mathbf{C_x}$.

If we wish to calculate the covariance matrix of vectors **y**, results

$$\mathbf{C_y} = E\{(\mathbf{y}\text{-}\mathbf{m_y})(\mathbf{y}\text{-}\mathbf{m_y})^T\} = E\{\mathbf{yy}^T\} \quad (7)$$

Because, $\mathbf{m_y}$ is a null vector. Besides, $\mathbf{C_y}$ is a diagonal matrix. Depending on the correlation degree between the original sub-blocks, KLT will be more or less efficient decorrelating them. Such efficiency depends on how the elements of the main diagonal of the covariance matrix $\mathbf{C_y}$ fall in value, from right to left. The faster they fall in value, the KLT will be more efficient decorrelating them. As an example, based on Fig.9, which represents to Lena of 512-by-512 pixels, and if we work with sub-blocks of 64-by-64 pixels, as we must see in Fig.10(a), we obtain the eigenvalues of Fig.10(b). However, if by a determined method we are starting from the a set of sub-blocks as those shown in Fig.10(c), then we will obtain the eigenvalues of Fig.10(d). The second case is highly more efficient than the first one.

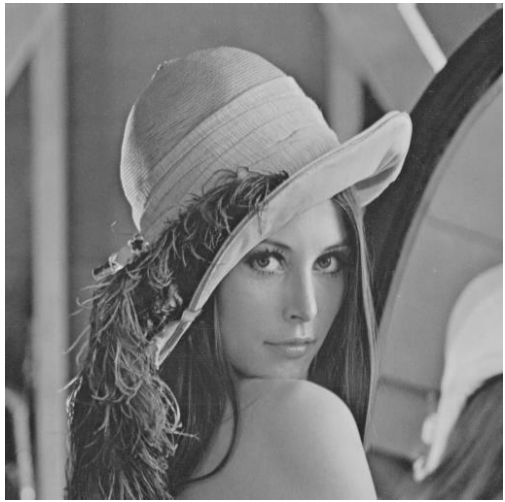

Fig. 9: Lena of 512-by-512 pixels, with 8 bits-per-pixel (bpp)

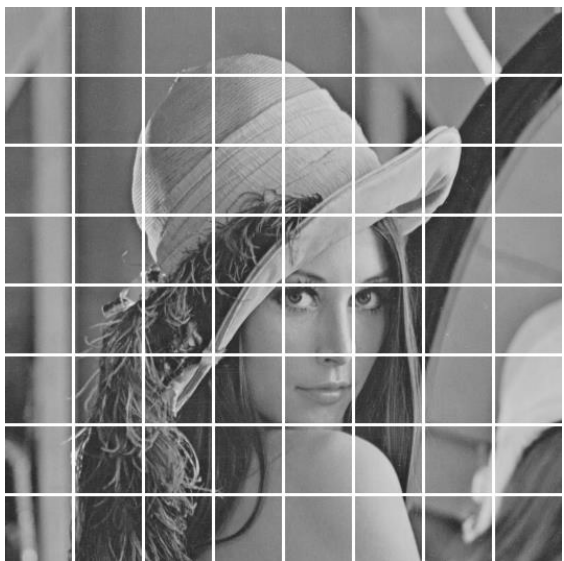

(a)

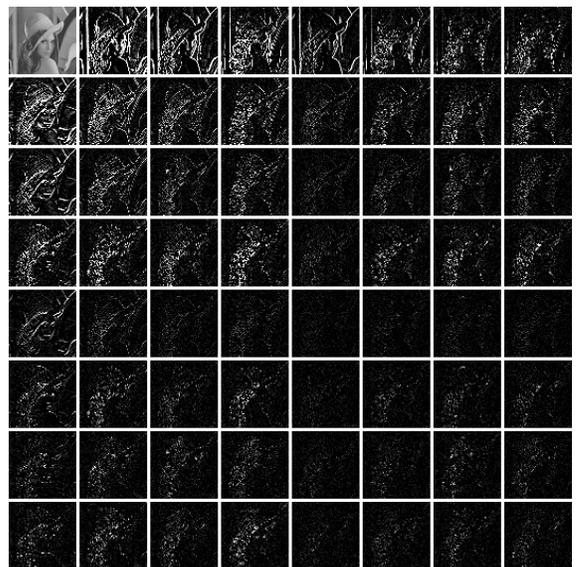

(c)

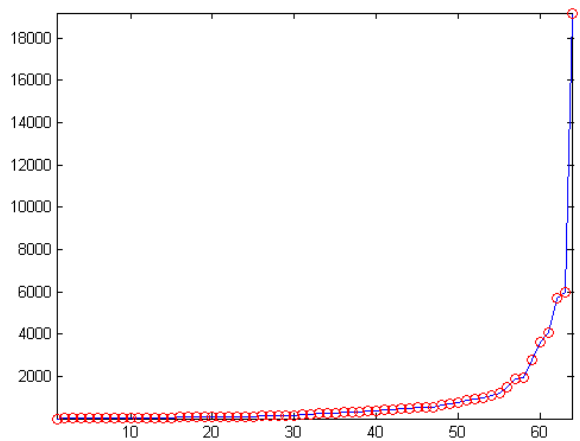

(b)

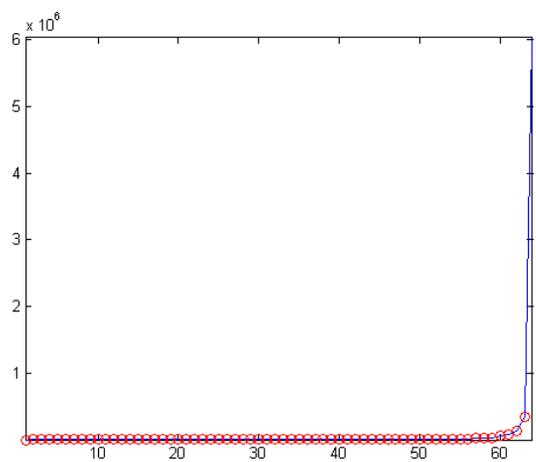

(d)

Fig. 10: Two different cases for efficiency evaluation, where: (a) original set of sub-blocks of 64-by-64 pixels, (b) they represents their eigenvalues, (c) efficient set of sub-blocks of 64-by-64 pixels, and (d) they represents their eigenvalues.

The Fig.10(c)-(d) represents a set of sub-blocks much more efficient than Fig.10(a)-(b), because, the sub-blocks of the Fig.10(c)-(d) are more correlated morphologically. In Fig.10(c) is evident than each sub-block represent a little version of Lena. In Fig.10(d) the last 2 sub-blocks account for about 95% of the total variance, while in Fig.10(a) the last 46 sub-blocks account for about 95% of the total variance. Therefore, Fig.10(a) is a inefficient set, while Fig.10(c) is highly efficient. This is the reason that makes the KLT as efficient in multi and hyperspectral imagery and very inefficient in images alone (monoframe) [1, 28-30, 32-34, 38-41]. A method prior to KLT (for monoframe images) which resulted in a high correlation of sub-blocks to make the KLT more efficient and will be very welcome.

On the other hands, the inverse KLT will be,

$$x = V\,y + m_x \quad (8)$$

A complete lossy image compression algorithm based on KLT may be:

**CODEC:**
1. Image sub-blocking with elected scan and construction of three dimensional matrix.
2. KLT to resulting sub-blocks
3. Pruning of sub-blocks based on percentage of resulting covariance matrix
4. Quantization
5. Entropy encoding

*To channel or storage*

**DECODEC:**
6. Entropy decoding
9. Complete with zeros the sub-blocks pruned
8. Inverse KLT
9. Reconstruction of bidimensional matrix from the new sub-blocks set with inverse scan and image reassembling.

## V. COMBINATIONS

Based on the last section, the proposed solutions to achieve the goal are as follows:

*First CODEC*
**1. Recursive Haar application to each sub-block depending on the final sub-blocks size**
**2. Morton's scan**
**3. Construction of three dimensional matrix**
**4. KLT**
**5. Pruning**
**6. Quantization**
**7. Entropy encoding**

*Second CODEC*
**1. Recursive Haar application to each sub-block depending on the final sub-blocks size**
**2. Row rafter scan**
**3. Construction of three dimensional matrix**
**4. KLT**
**5. Pruning**
**6. Quantization**
**7. Entropy encoding**

However, starting in both codecs, What the meaning of "*Recursive Haar application to each sub-block depending on the final sub-blocks size*"? If we call **J** to the mentioned three dimensional matrix (see Line 3 of both CODECs), and based on Fig.11 for the first level both scans match.

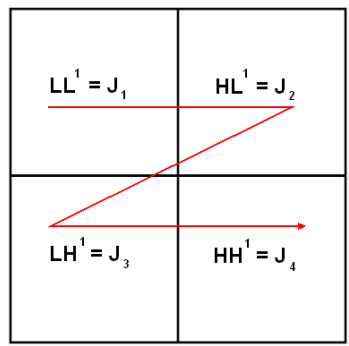

Fig. 11: Morton's and row rafter scanning

But, for the second level of Haar applications the used scan generates two different three dimensional matrices J. See Fig.12 for Morton's scan and Fig.13 for row rafter scan.

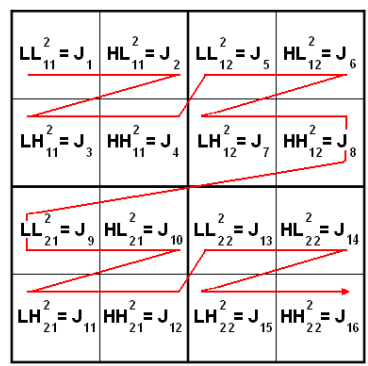

Fig. 12: Morton's scanning

The key of both scans and the subsequent formation of the J matrix are its subscript. While, the superscript of LL, LH, HL and HH Haar's sub-bands represent the level of application of the DWT.

Finally, the difference between the two will be (with and without noise) through simulations of Section VII.

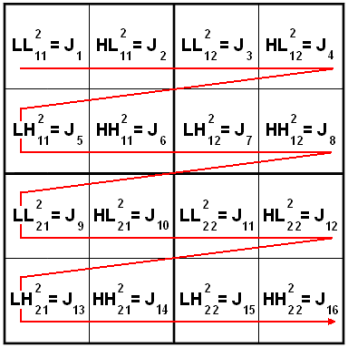

Fig. 13: Row rafter scanning

## VI. METRICS

### A. Data Compression Ratio (CR)

Data compression ratio, also known as compression power, is a computer-science term used to quantify the reduction in data-representation size produced by a data compression algorithm. The data compression ratio is analogous to the physical compression ratio used to measure physical compression of substances, and is defined in the same way, as the ratio between the *uncompressed size* and the *compressed size* [59]:

$$CR = \frac{Uncompressed\ Size}{Compressed\ Size} \quad (9)$$

Thus a representation that compresses a 10MB file to 2MB has a compression ratio of 10/2 = 5, often notated as an explicit ratio, 5:1 (read "five to one"), or as an implicit ratio, 5X. Note that this formulation applies equally for compression, where the uncompressed size is that of the original; and for decompression, where the uncompressed size is that of the reproduction.

### B. Peak Signal-To-Noise Ratio (PSNR)

The phrase peak signal-to-noise ratio, often abbreviated PSNR, is an engineering term for the ratio between the maximum possible power of a signal and the power of corrupting noise that affects the fidelity of its representation. Because many signals have a very wide dynamic range, PSNR is usually expressed in terms of the logarithmic decibel scale.

The PSNR is most commonly used as a measure of quality of reconstruction in image compression, etc [59]. It is most easily defined via the mean squared error (MSE) which for two $NR \times NC$ (rows-by-columns) monochrome images $I$ and $I_d$, where the second one of the images is considered a denoised approximation of the other is defined as:

$$MSE = \frac{1}{NRxNC} \sum_{nr=0}^{NR-1} \sum_{nc=0}^{NC-1} \left\| I(nr,nc) - I_d(nr,nc) \right\|^2 \quad (10)$$

The PSNR is defined as [59]:

$$PSNR = 10 \log_{10}(\frac{MAX_I^2}{MSE}) = 20 \log_{10}(\frac{MAX_I}{\sqrt{MSE}}) \quad (11)$$

Here, $MAX_I$ is the maximum pixel value of the image. When the pixels are represented using 8 bits per sample, this is 256. More generally, when samples are represented using linear pulse code modulation (PCM) with B bits per sample, maximum possible value of $MAX_I$ is $2^B-1$.

For color images with three red-green-blue (RGB) values per pixel, the definition of PSNR is the same except the MSE is the sum over all squared value differences divided by image size and by three [59].

Typical values for the PSNR in lossy image and video compression are between 30 and 50 dB, where higher is better.

## VII. COMPUTERS SIMULATIONS

The simulations are organized in four experiments. In all cases where we used DWT (Haar's basis), we used soft-thresholding for high sub-bands coefficients shrinkage. In each experiment, we are going to compare seven different compression techniques:

1. DWT according to the Section II.
2. DWT plus Morton's scan.
3. DWT plus row-rafter scan.
4. Morton's scan plus KLT.
5. Row-rafter scan plus KLT.
6. DWT plus Morton's scan plus KLT.
7. DWT plus row-rafter scan plus KLT.

All experiments include calculations of MSE and PSNR.

### A. Experiment 1:
Main characteristics:
1. Image = Lena
2. Color = gray
3. Size = 512-by-512 pixels
4. Bits-per-pixel = 8
5. Maximum compression rate = 4:1
6. Sub-blocks size = 64-by-64 pixels for compression techniques 4, 5, 6 and 7.
    = 256-by-256 pixels for compression techniques 2 and 3.
7. Level of decomposition for DWT alone = 1
8. Noisy = no

Fig.14 shows the seven techniques with original image, while, Fig.15 shows the sub-blocks set of the compression techniques 4, 5, 6 and 7 with their respective eigenvalues distributions. Table I shoes the metrics (MSE, PSNR and CR) for Fig.14.

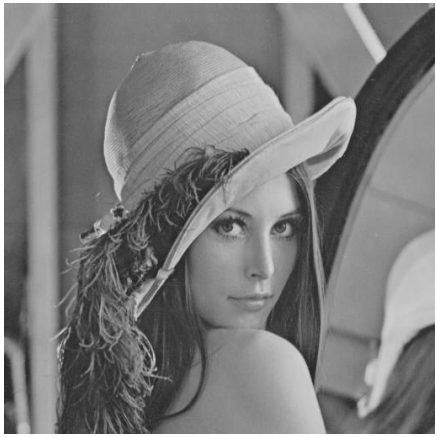
(a)
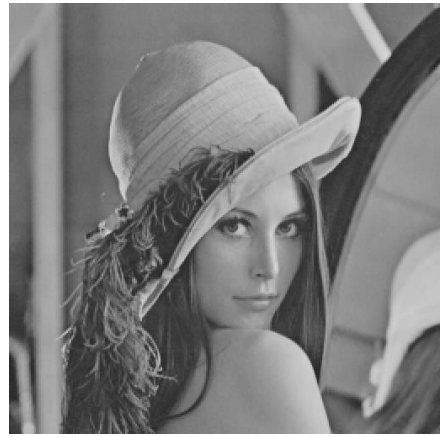
(b)
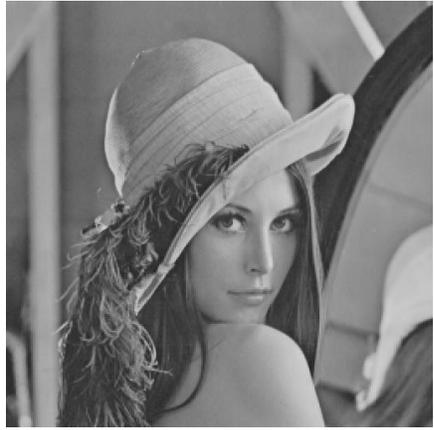
(c)
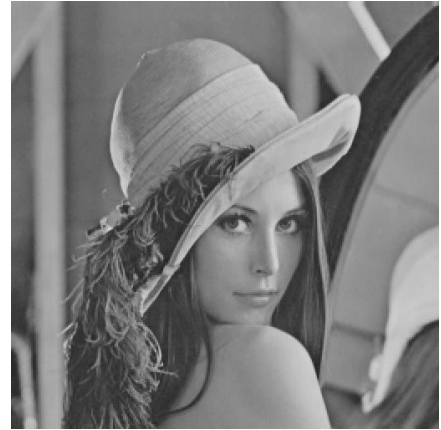
(d)
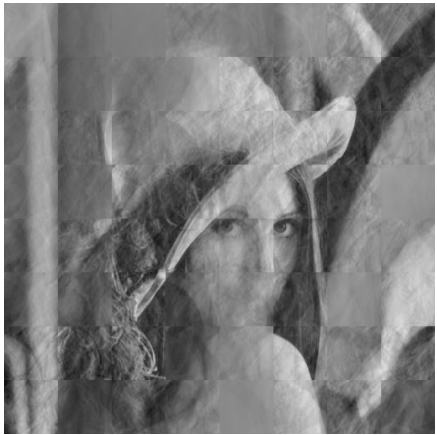
(e)
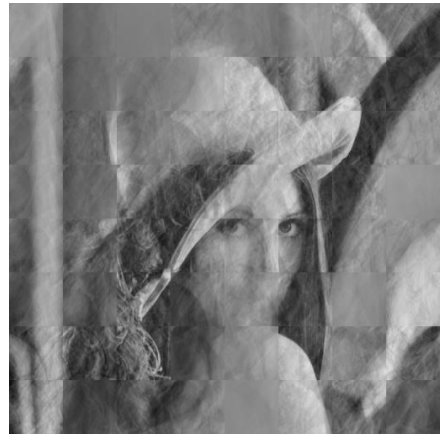
(f)
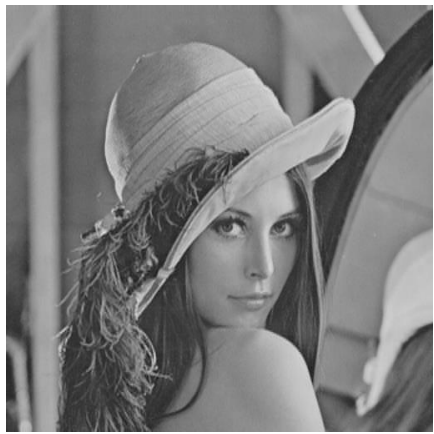
(g)
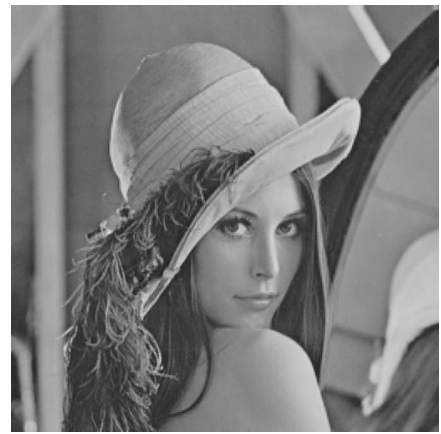
(h)

Fig.14: Comparisson of compression techniques: (a) original, (b) Haar, (c) Haar + Morton's scan, (d) Haar + row-rafter scan, (e) Morton's scan + KLT, (f) row-rafter scan + KLT, (g) Haar + Morton's scan + KLT, and (h) Haar + row-rafter scan + KLT

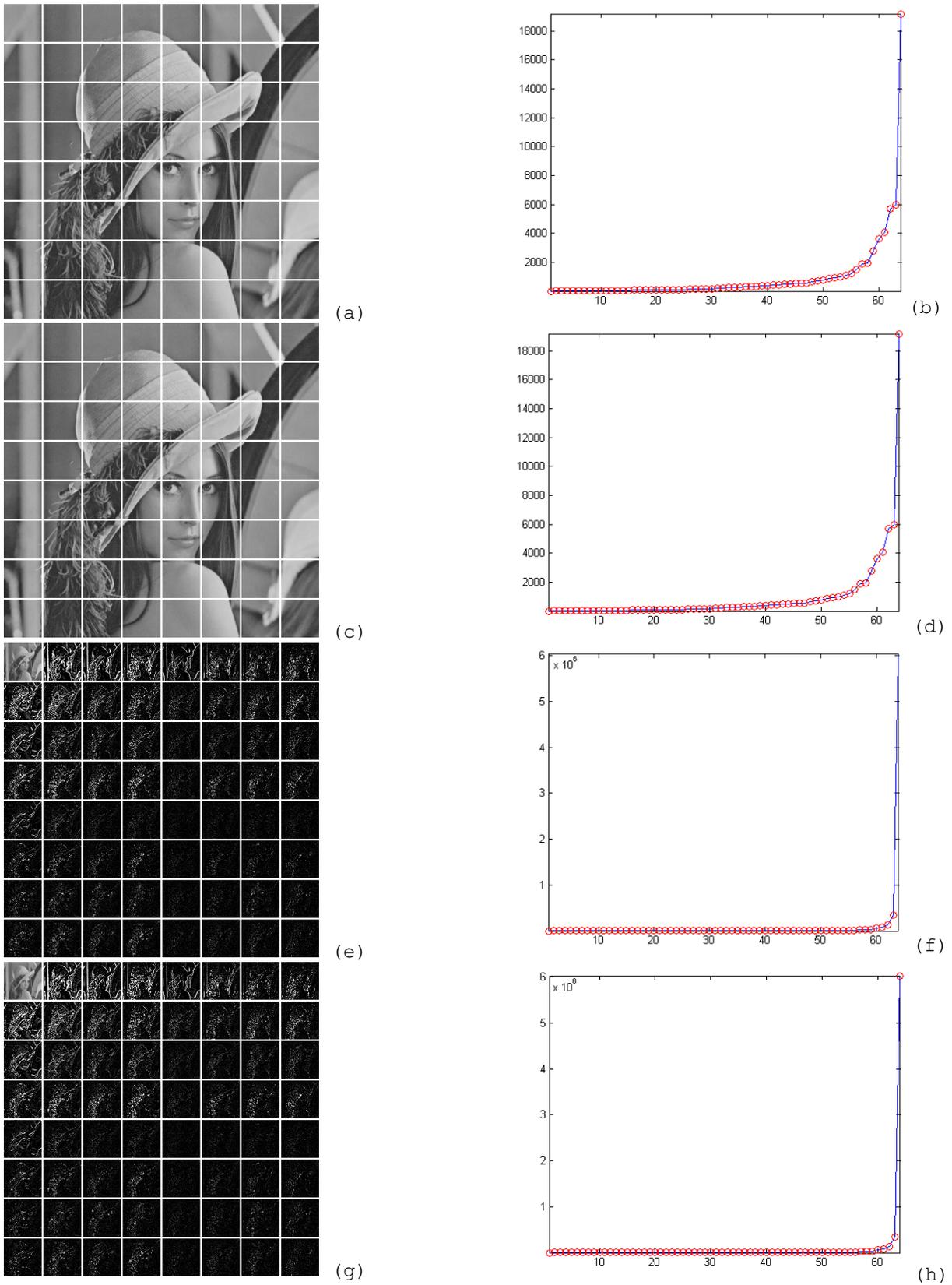

Fig.15 Comparisson of efficiency: (a) sub-blocks set for Morton's scan + KLT, (b) eigenvalues distribution for Morton's scan + KLT, (c) sub-blocks set for row-rafter scan + KLT, (d) eigenvalues distribution for row-rafter scan + KLT, (e) sub-blocks set for Haar + Morton's scan + KLT, (f) eigenvalues distribution for Haar + Morton's scan + KLT, (g) sub-blocks set for Haar + row-rafter scan + KLT, and (h) eigenvalues distribution for Haar + row-rafter scan + KLT

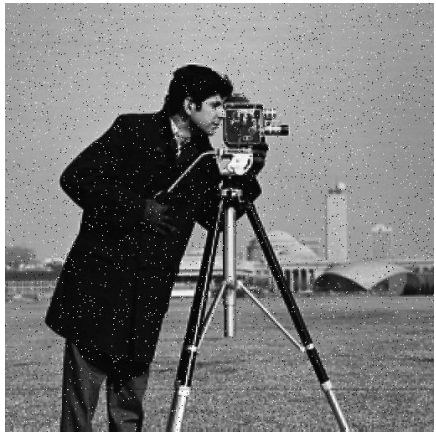
(a)
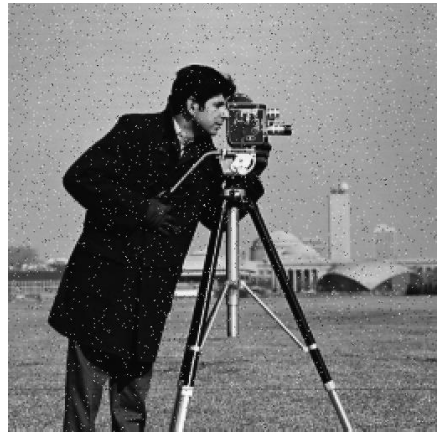
(b)
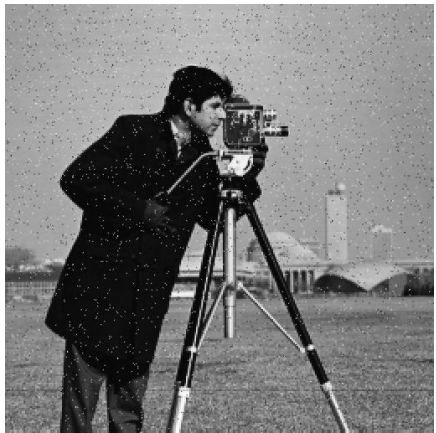
(c)
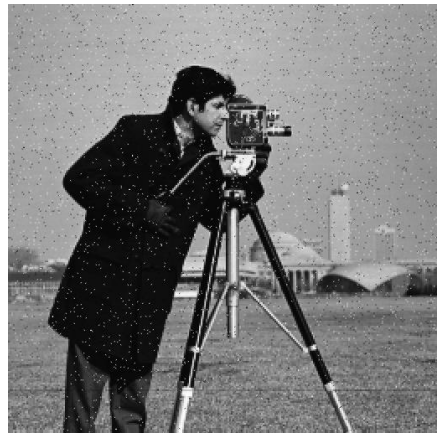
(d)
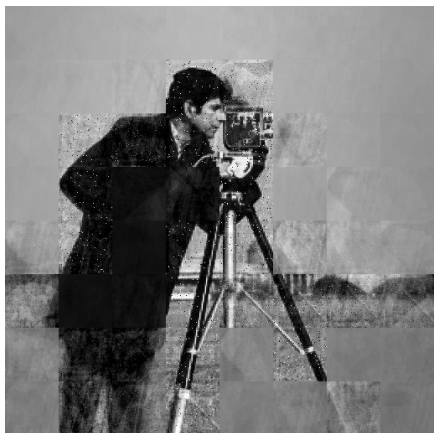
(e)
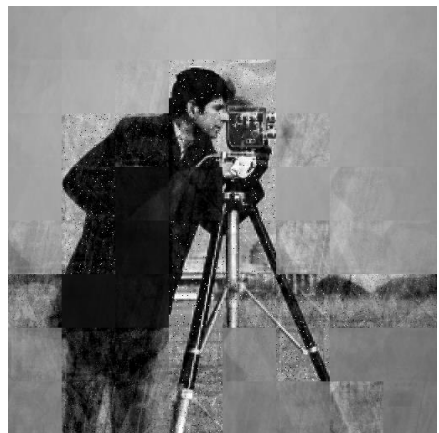
(f)
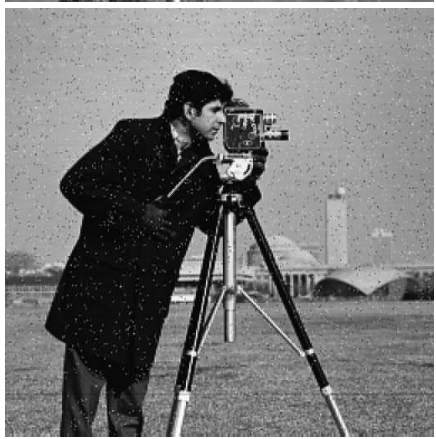
(g)
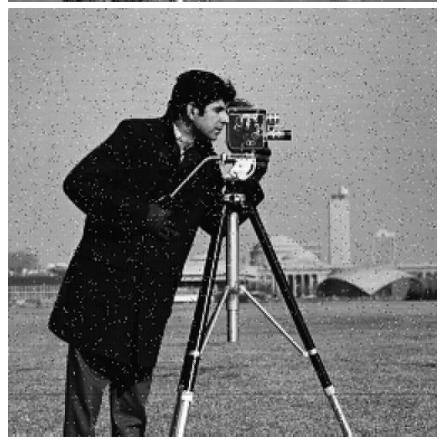
(h)

Fig.16: Comparisson of compression techniques: (a) original, (b) Haar, (c) Haar + Morton's scan, (d) Haar + row-rafter scan, (e) Morton's scan + KLT, (f) row-rafter scan + KLT, (g) Haar + Morton's scan + KLT, and (h) Haar + row-rafter scan + KLT

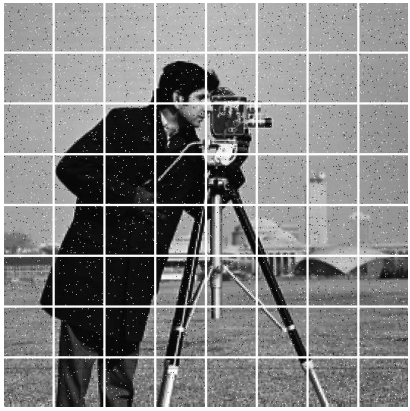 (a)
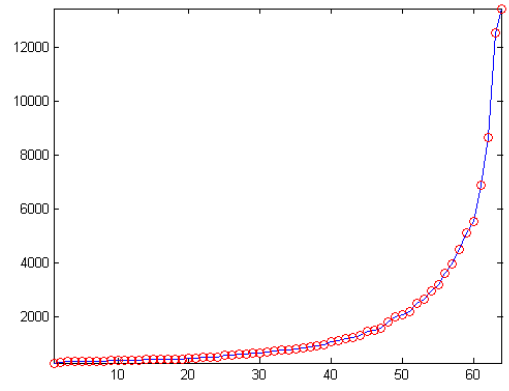 (b)
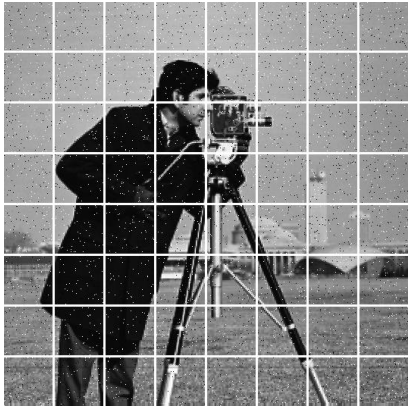 (c)
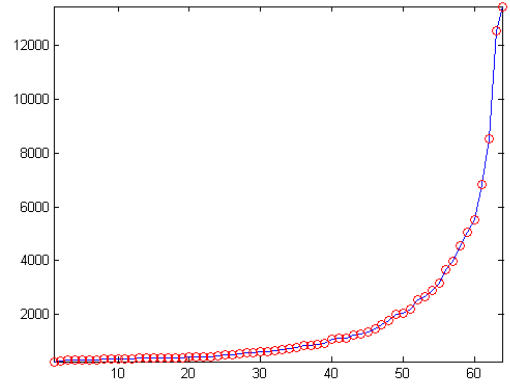 (d)
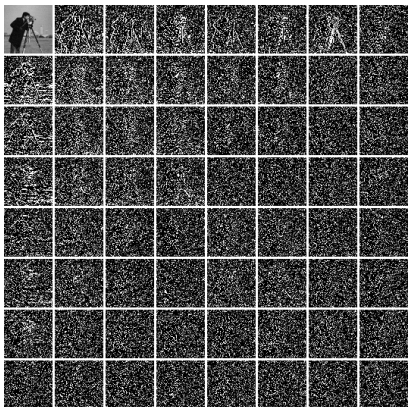 (e)
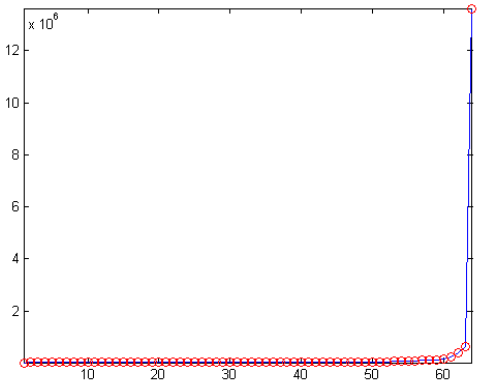 (f)
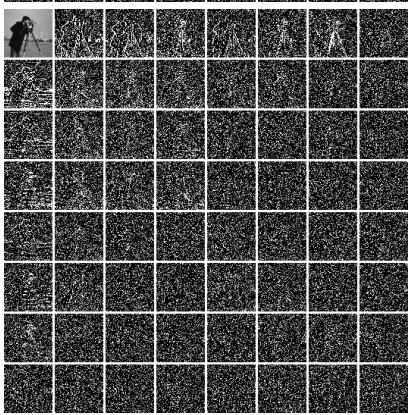 (g)
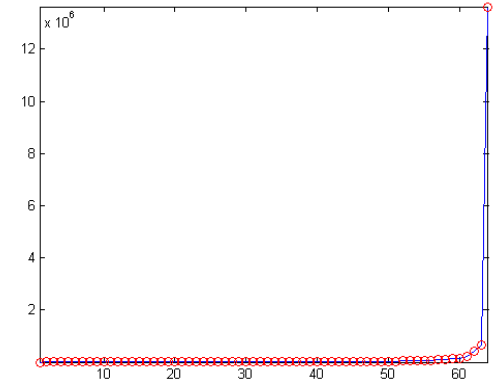 (h)

Fig.17 Comparisson of efficiency: (a) sub-blocks set for Morton's scan + KLT, (b) eigenvalues distribution for Morton's scan + KLT, (c) sub-blocks set for row-rafter scan + KLT, (d) eigenvalues distribution for row-rafter scan + KLT, (e) sub-blocks set for Haar + Morton's scan + KLT, (f) eigenvalues distribution for Haar + Morton's scan + KLT, (g) sub-blocks set for Haar + row-rafter scan + KLT, and (h) eigenvalues distribution for Haar + row-rafter scan + KLT

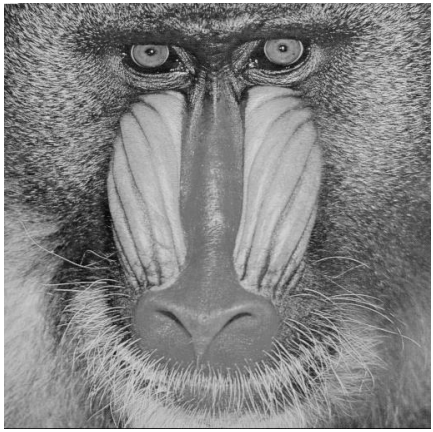
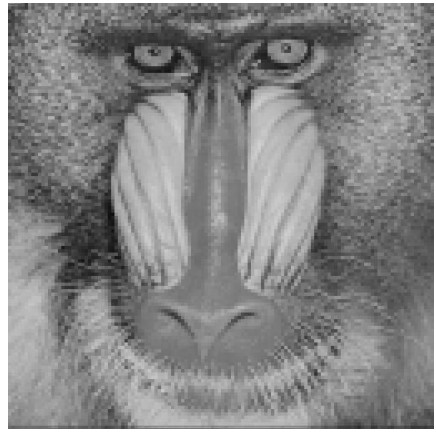
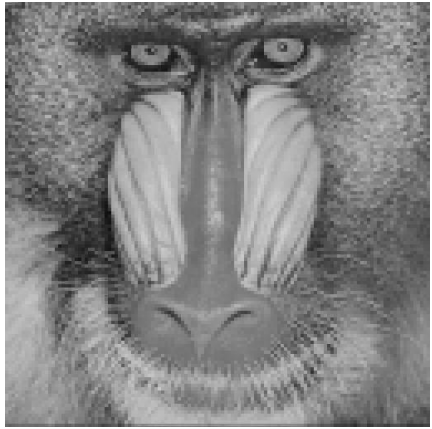
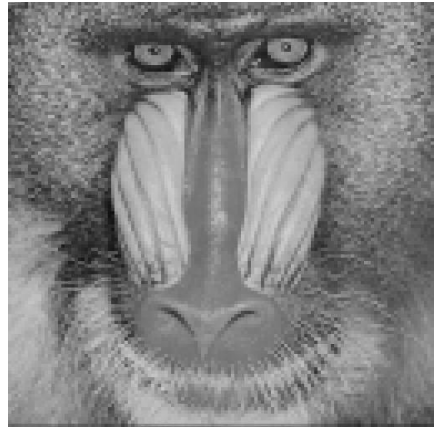
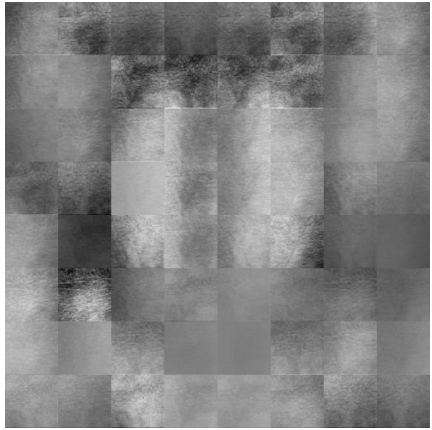
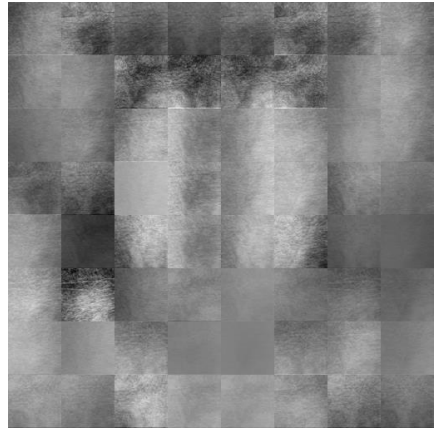
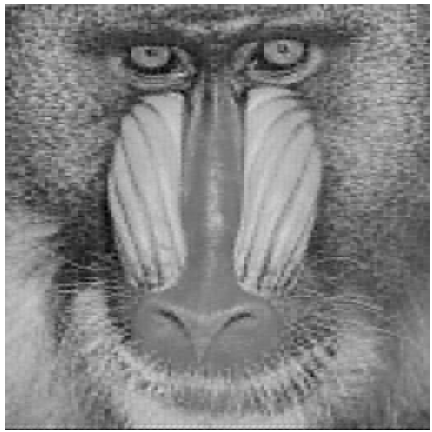
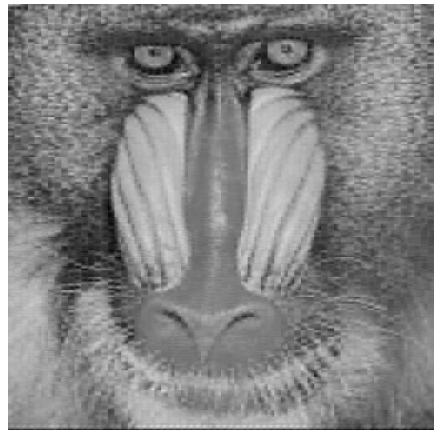

Fig.18: Comparisson of compression techniques: (a) original, (b) Haar, (c) Haar + Morton's scan, (d) Haar + row-rafter scan, (e) Morton's scan + KLT, (f) row-rafter scan + KLT, (g) Haar + Morton's scan + KLT, and (h) Haar + row-rafter scan + KLT

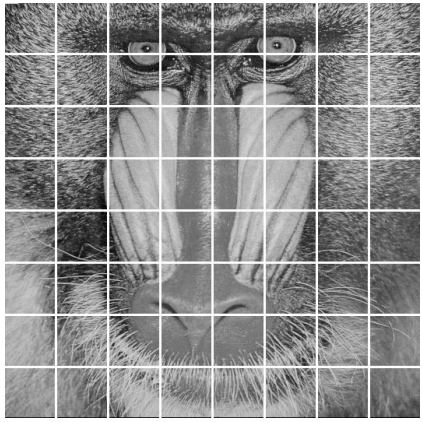
(a)
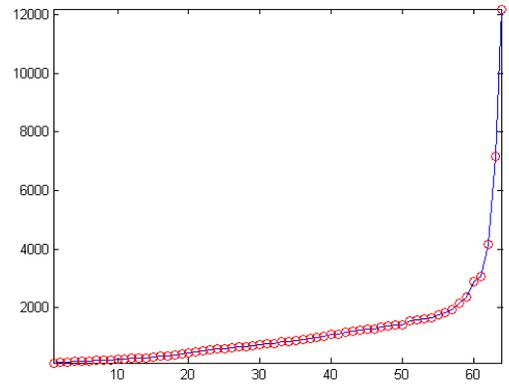
(b)
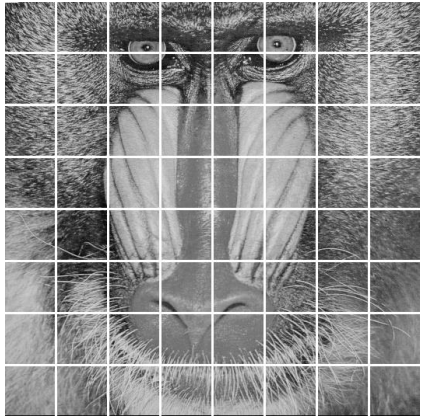
(c)
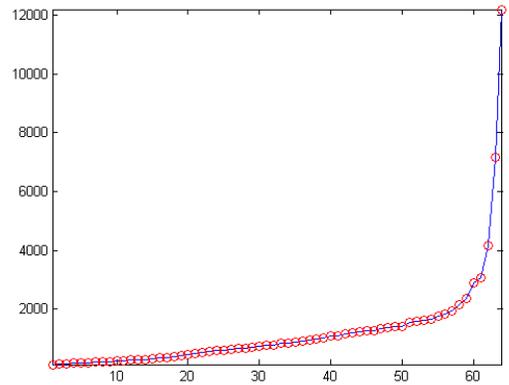
(d)
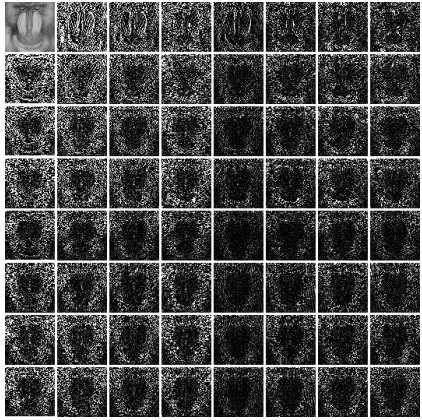
(e)
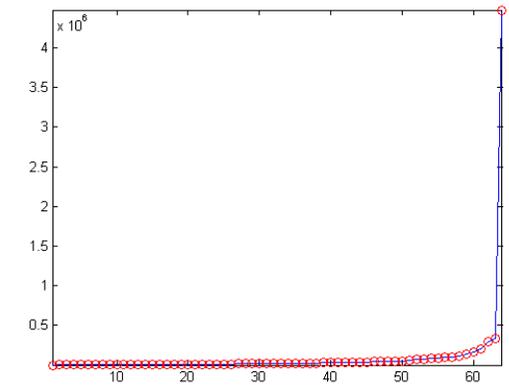
(f)
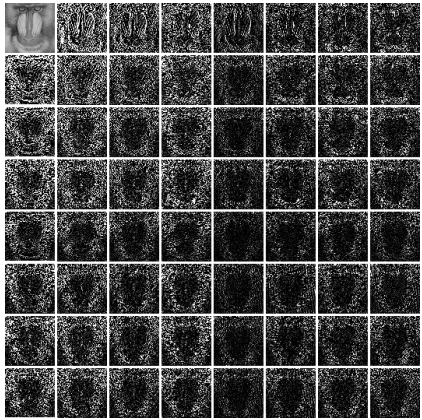
(g)
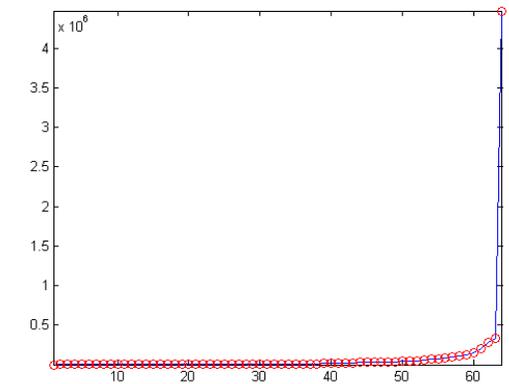
(h)

Fig.19 Comparisson of efficiency: (a) sub-blocks set for Morton's scan + KLT, (b) eigenvalues distribution for Morton's scan + KLT, (c) sub-blocks set for row-rafter scan + KLT, (d) eigenvalues distribution for row-rafter scan + KLT, (e) sub-blocks set for Haar + Morton's scan + KLT, (f) eigenvalues distribution for Haar + Morton's scan + KLT, (g) sub-blocks set for Haar + row-rafter scan + KLT, and (h) eigenvalues distribution for Haar + row-rafter scan + KLT

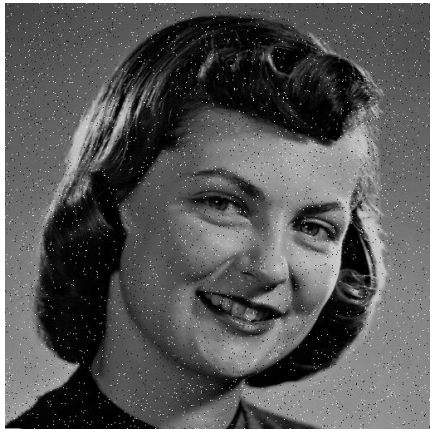
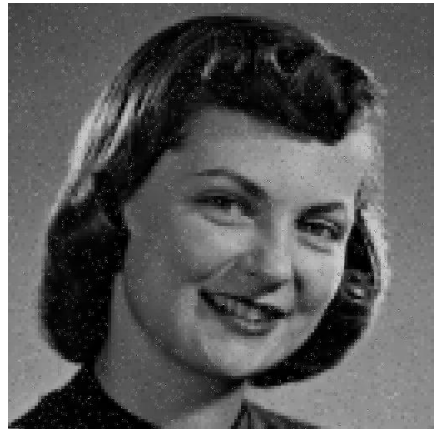
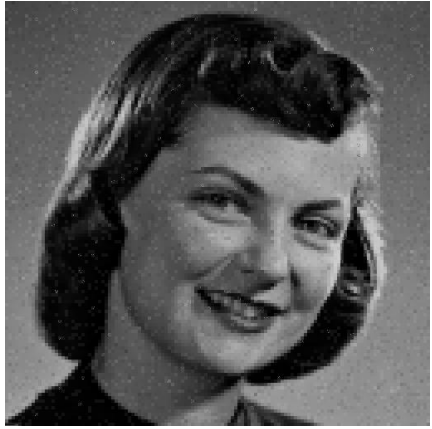
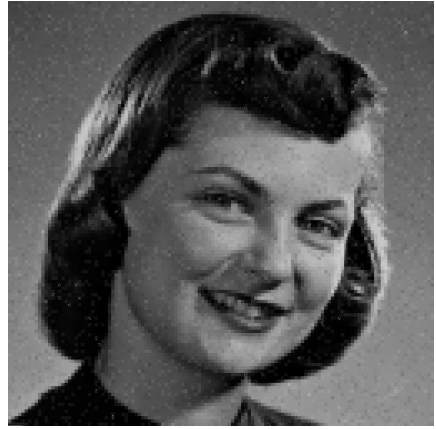
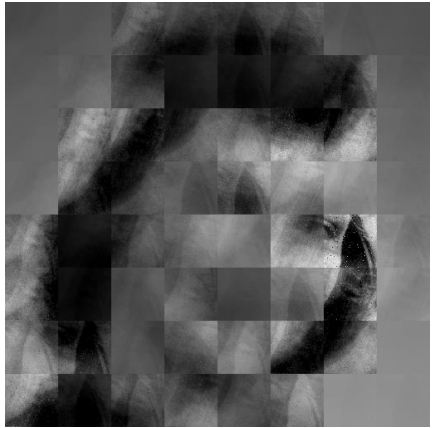
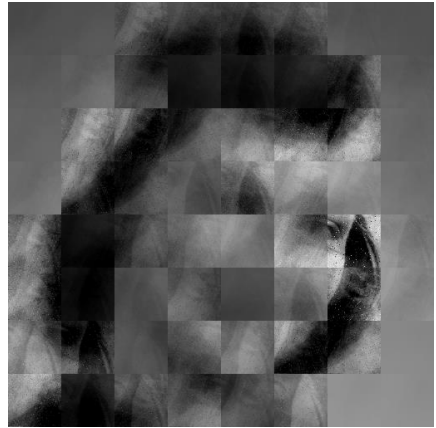
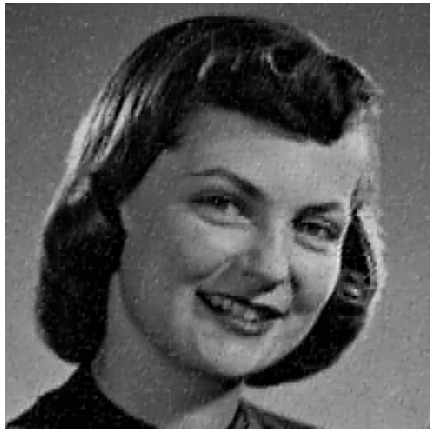
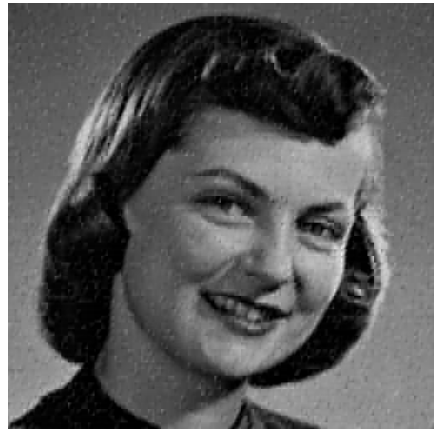

Fig.20: Comparisson of compression techniques: (a) original, (b) Haar, (c) Haar + Morton's scan, (d) Haar + row-rafter scan, (e) Morton's scan + KLT, (f) row-rafter scan + KLT, (g) Haar + Morton's scan + KLT, and (h) Haar + row-rafter scan + KLT

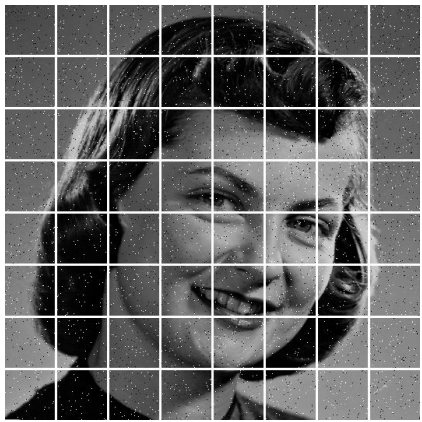 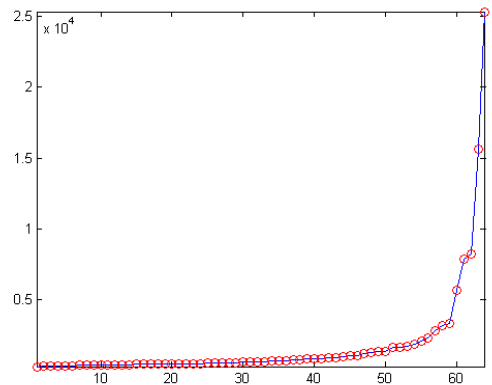
(a) (b)
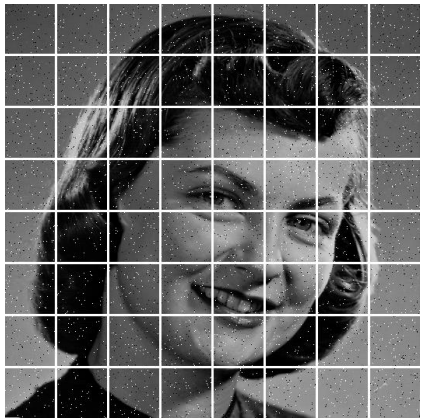 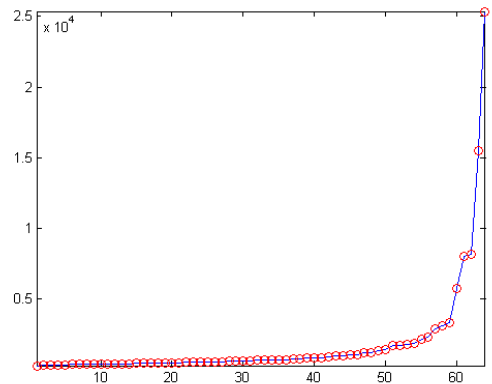
(c) (d)
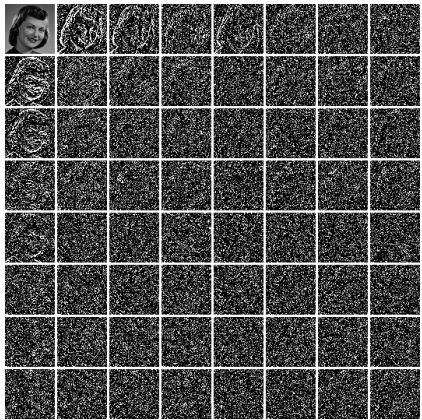 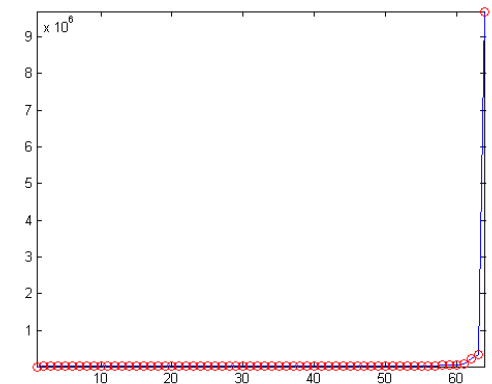
(e) (f)
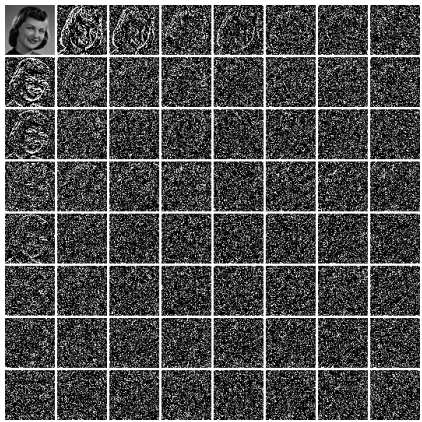 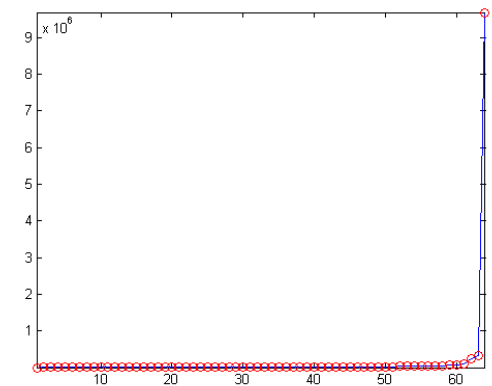
(g) (h)

Fig.21 Comparisson of efficiency: (a) sub-blocks set for Morton's scan + KLT, (b) eigenvalues distribution for Morton's scan + KLT, (c) sub-blocks set for row-rafter scan + KLT, (d) eigenvalues distribution for row-rafter scan + KLT, (e) sub-blocks set for Haar + Morton's scan + KLT, (f) eigenvalues distribution for Haar + Morton's scan + KLT, (g) sub-blocks set for Haar + row-rafter scan + KLT, and (h) eigenvalues distribution for Haar + row-rafter scan + KLT

TABLE I
Experiment 1: Lena, CR = 4:1

| Technique | CR | MSE | PSNR |
|---|---|---|---|
| Haar | 3.8790 | 26.9291 | 33.8286 |
| Haar+Morton | 3.8790 | 26.9291 | 33.8286 |
| Haar+row-rafter | 3.8790 | 26.9291 | 33.8286 |
| Morton+KLT | 3.9347 | 138.3011 | 26.7225 |
| Row-rafter+KLT | 3.9347 | 138.3011 | 26.7225 |
| Haar+Morton+KLT | 3.9347 | 13.5447 | 36.8131 |
| Haar+row-rafter+KLT | 3.9347 | 13.5447 | 36.8131 |

*B. Experiment 2:*
Main characteristics:
1. Image = Camera
2. Color = gray
3. Size = 512-by-512 pixels
4. Bits-per-pixel = 8
5. Maximum compression rate = 4:1
6. Sub-blocks size = 64-by-64 pixels for compression techniques 4, 5, 6 and 7.
   = 256-by-256 pixels for compression techniques 2 and 3.
7. Level of decomposition for DWT alone = 1
8. Noisy = type SALT and pepper, mean = 0, and variante = 0.02

Fig.16 shows the seven techniques with original image, while, Fig.17 shows the sub-blocks set of the compression techniques 4, 5, 6 and 7 with their respective eigenvalues distributions. Table II shoes the metrics (MSE, PSNR and CR) for Fig.16.

TABLE II
Experiment 2: Camera, CR = 4:1

| Technique | CR | MSE | PSNR |
|---|---|---|---|
| Haar | 3.4853 | 173.7846 | 25.7307 |
| Haar+Morton | 3.4811 | 170.8111 | 25.8056 |
| Haar+row-rafter | 3.4740 | 170.0747 | 25.8244 |
| Morton+KLT | 3.9347 | 491.1751 | 21.2184 |
| Row-rafter+KLT | 3.9347 | 487.7782 | 21.2486 |
| Haar+Morton+KLT | 3.9347 | 308.9114 | 23.2325 |
| Haar+row-rafter+KLT | 3.9347 | 310.5899 | 23.2089 |

*C. Experiment 3:*
Main characteristics:
1. Image = Baboon
2. Color = gray
3. Size = 512-by-512 pixels
4. Bits-per-pixel = 8
5. Maximum compression rate = 16:1
6. Sub-blocks size = 64-by-64 pixels for compression techniques 4, 5, 6 and 7.
   = 128-by-128 pixels for compression techniques 2 and 3.
7. Level of decomposition for DWT alone = 2
8. Noisy = no

Fig.18 shows the seven techniques with original image, while, Fig.19 shows the sub-blocks set of the compression techniques 4, 5, 6 and 7 with their respective eigenvalues distributions. Table III shoes the metrics (MSE, PSNR and CR) for Fig.18.

TABLE III
Experiment 3: Baboon, CR = 16:1

| Technique | CR | MSE | PSNR |
|---|---|---|---|
| Haar | 15.6336 | 526.0533 | 20.9205 |
| Haar+Morton | 15.4103 | 524.7629 | 20.9312 |
| Haar+row-rafter | 15.4103 | 524.7629 | 20.9312 |
| Morton+KLT | 15.6935 | 818.7159 | 18.9995 |
| Row-rafter+KLT | 15.6935 | 818.7159 | 18.9995 |
| Haar+Morton+KLT | 15.6935 | 489.8204 | 21.2304 |
| Haar+row-rafter+KLT | 15.6935 | 489.8204 | 21.2304 |

*D. Experiment 4:*
Main characteristics:
1. Image = Girl
2. Color = gray
3. Size = 512-by-512 pixels
4. Bits-per-pixel = 8
5. Maximum compression rate = 16:1
6. Sub-blocks size = 64-by-64 pixels for compression techniques 4, 5, 6 and 7.
   = 128-by-128 pixels for compression techniques 2 and 3.
7. Level of decomposition for DWT alone = 2
8. Noisy = type SALT and pepper, mean = 0, and variante = 0.02

Fig.20 shows the seven techniques with original image, while, Fig.12 shows the sub-blocks set of the compression techniques 4, 5, 6 and 7 with their respective eigenvalues distributions. Table IV shoes the metrics (MSE, PSNR and CR) for Fig.20.

TABLE IV
Experiment 4: Girl, CR = 16:1

| Technique | CR | MSE | PSNR |
|---|---|---|---|
| Haar | 10.2376 | 308.1543 | 23.2431 |
| Haar+Morton | 14.8238 | 467.2525 | 21.4353 |
| Haar+row-rafter | 14.8439 | 464.3150 | 21.4627 |
| Morton+KLT | 15.6935 | 886.9586 | 18.6518 |
| Row-rafter+KLT | 15.6935 | 881.5347 | 18.6784 |
| Haar+Morton+KLT | 15.6935 | 443.9665 | 21.6573 |
| Haar+row-rafter+KLT | 15.6935 | 460.9388 | 21.4944 |

Finally, all techniques (denoising and compression) were implemented in MATLAB® (Mathworks, Natick, MA) [60] on a PC with an Intel® Core(TM) QUAD CPU Q6600 2.40 GHz processors and 4 GB RAM.

## VIII. CONCLUSION

A first and relevant clarification is as follows, in theory, KLT ordered their eigenvalues from highest to lowest [28]. However, as we explained in the previous section, we use MATLAB® in all our simulations, in particular the built-in function "eigen". Formally, we use it as follows [$\mathbf{V}$,$\mathbf{C_y}$] = eigen($\mathbf{C_x}$), to obtain matrices $\mathbf{V}$ and $\mathbf{C_y}$. Where $\mathbf{C_y}$ is a diagonal matrix, where in its main diagonal has the eigenvalues of $\mathbf{C_x}$. Such eigenvalues are ordered from lowest to highest [60].

This does not change anything in the calculations, if the consistent order is respected for all the variables involved in the problem context.

As shown in the Figures 15, 17, 19 and 21, although KLT is optimum, it is inefficient in the sub-blocks decorrelation, in the cases where such sub-blocks are morphologically differents. The experimental evidence shows that previous DWT supplies KLT of the necessary morphological affinity, see Figures 14, 16, 18 and 20.

*Experiment 1:*
Haar, Haar+Morton and Haar+row-rafter have identical metric values. Morton+KLT and row-rafter+KLT have identical metric values. Haar+Morton+KLT and Haar+row-rafter+KLT have identical metric values too.

*Experiment 2:*
The noise causes different metric values in all. However, Morton+KLT, row-rafter+KLT, Haar+Morton+KLT and Haar+row-rafter+KLT have similar real CR.

*Experiment 3:*
Similar situation to Experiment 1.
In Morton+KLT and row-rafter+KLT the block effect is obvious, and they have a very bad look-and-feel, that is to say, image quality.

*Experiment 4:*
Similar situation to Experiment 2.
In Morton+KLT and row-rafter+KLT the block effect is obvious too, with similar consequences.

In the four experiments Haar+Morton+KLT is better than Morton+KLT, and Haar+row-rafter+KLT is better than row-rafter+KLT.

On the other hands, without noise the scan type results trivial.

As discussed earlier, the KLT is theoretically the optimum method to spectrally decorrelate a set of sub-blocks image. However, it is computationally expensive. Future research should be geared to the use of lower-cost computational approaches [61-63].

ACKNOWLEDGMENT

M. Mastriani thanks Prof. Marta Mejail, director of Laboratorio de Visión Robótica y Procesamiento de Imágenes del Departamento de Computación de la Facultad de Ciencias Exactas y Naturales de la Universidad de Buenos Aires, for her tremendous help and support.

REFERENCES

[1] J.-L. Starck, and P. Querre, "Multispectral Data Restoration by the Wavelet-Karhunen-Loève Transform," Preprint submitted to Elsevier Preprint, 2000, pp. 1–29.
[2] D. L. Donoho, and I. M. Johnstone, "Adapting to unknown smoothness via wavelet shrinkage," Journal of the American Statistical Assoc., vol. 90, no. 432, pp. 1200-1224., 1995.
[3] D. L. Donoho, and I. M. Johnstone, "Ideal spatial adaptation by wavelet shrinkage," Biometrika, 81, 425-455, 1994.
[4] I. Daubechies. Ten Lectures on Wavelets, SIAM, Philadelphia, PA. 1992.
[5] I. Daubechies, "Different Perspectives on Wavelets," in *Proceedings of Symposia in Applied Mathematics*, vol. 47, American Mathematical Society, USA, 1993.
[6] S. Mallat, "A theory for multiresolution signal decomposition: The wavelet representation," *IEEE Trans. Pattern Anal. Machine Intell.*, vol. 11, pp. 674–693, July 1989.
[7] S. G. Mallat, "Multiresolution approximations and wavelet orthonormal bases of L2 (R)," *Transactions of the American Mathematical Society*, 315(1), pp.69-87, 1989a.
[8] S. G. Chang, B. Yu, and M. Vetterli, "Adaptive wavelet thresholding for image denoising and compression," *IEEE Trans. Image Processing*, vol. 9, pp. 1532–1546, Sept. 2000.
[9] M. Misiti, Y. Misiti, G. Oppenheim, and J.M. Poggi. (2001, June). Wavelet Toolbox, for use with MATLAB®, User's guide, version 2.1.
[10] C.S. Burrus, R.A. Gopinath, and H. Guo, *Introduction to Wavelets and Wavelet Transforms: A Primer*, Prentice Hall, New Jersey, 1998.
[11] B.B. Hubbard, *The World According to Wavelets: The Story of a Mathematical Technique in the Making*, A. K. Peter Wellesley, Massachusetts, 1996.
[12] A. Grossman and J. Morlet, "Decomposition of Hardy Functions into Square Integrable Wavelets of Constant Shape," *SIAM J. App Math*, 15: pp.723-736, 1984.
[13] C. Valens. (2004). A really friendly guide to wavelets.
[14] G. Kaiser, *A Friendly Guide To Wavelets*, Boston: Birkhauser, 1994.
[15] J.S. Walker, *A Primer on Wavelets and their Scientific Applications*, Chapman & Hall/CRC, New York, 1999.
[16] E. J. Stollnitz, T. D. DeRose, and D. H. Salesin, *Wavelets for Computer Graphics: Theory and Applications*, Morgan Kaufmann Publishers, San Francisco, 1996.
[17] J. Shen and G. Strang, "The zeros of the Daubechies polynomials," in *Proc. American Mathematical Society,* 1996.
[18] H.Y. Gao, and A.G. Bruce, "WaveShrink with firm shrinkage," *Statistica Sinica*, 7, 855-874, 1997.
[19] V.S. Shingate, *et al*, "Still image compression using Embedded Zerotree Wavelet Encoding," In *International Conference on Cognitive Systems* New Delhi, December 14-15, pp.1-9, 2004.
[20] C. Valens, EZW encoding, 2004.
[21] Yuan-Yuan HU, *et al*, "Embedded Wavelet Image Compression Algorithm Based on Full Zerotree," In IJCSES International Journal of Computer Sciences and Engineering Systems, Vol.1, No.2, pp. 131-138, April 2007.
[22] C.Wang, and K-L. Ma, "A Statistical Approach to Volume Data Quality Assessment," IEEE Transactions on Visualization and Computer Graphics, Vol.14, No. 3, May/June, 2008, pp. 590-602.
[23] R. Muller, A study of image compression techniques, with specific focus on weighted finite automata, Thesis for Degree Master of Science at the University of Stellenbosch, December 2005.
[24] S.S. Polisetty, Hardware acceleration of the embedded zerotree wavelet algorithm, Thesis for Master of Science Degree at the University of Tennessee, Knoxville, December 2004.
[25] V. Mahomed, and S.H. Mneney, "Wavelet based compression: the new still image compression technique," In Sixteenth Annual Symposium of the Pattern Recognition Association of South Africa, 23-25 November 2005, Langebaan, South Africa, pp.67-72.
[26] H.A. Kim Taavo, Scalable video using wavelets, Master of Science Programme, Lulea University of Technology, May 14, 2002.
[27] D. Zhang, and S. Chen, "Fast image compression using matrix K-L transform," Neurocomputing, Volume 68, pp.258-266, October 2005.
[28] R.C. Gonzalez, R.E. Woods, Digital Image Processing, 2nd Edition, Prentice- Hall, Jan. 2002, pp.675-683.
[29] -, The transform and data compression handbook, Edited by K.R. Rao, and P.C. Yip, CRC Press Inc., Boca Raton, FL, USA, 2001.
[30] B.R. Epstein, *et al*, "Multispectral KLT-wavelet data compression for landsat thematic mapper images," In *Data Compression Conference*, pp. 200-208, Snowbird, UT, March 1992.

[31] K. Konstantinides, *et al*, "Noise using block-based singular value decomposition," *IEEE Transactions on Image Processing*, 6(3), pp.479-483, 1997.

[32] J. Lee, "Optimized quadtree for Karhunen-Loève Transform in multispectral image coding," *IEEE Transactions on Image Processing*, 8(4), pp.453-461, 1999.

[33] J.A. Saghri, *et al*, "Practical Transform coding of multispectral imagery," *IEEE Signal Processing Magazine*, 12, pp.32-43, 1995.

[34] T-S. Kim, *et al*, "Multispectral image data compression using classified prediction and KLT in wavelet transform domain," IEICE Transactions on Fundam Electron Commun Comput Sci, Vol. E86-A; No.6, pp.1492-1497, 2003.

[35] J.L. Semmlow, Biosignal and biomedical image processing: MATLAB-Based applications, Marcel Dekker, Inc., New York, 2004.

[36] S. Borman, and R. Stevenson, "Image sequence processing," Department, Ed. Marcel Dekker, New York, 2003. pp.840-879.

[37] M. Wien, Variable Block-Size Transforms for Hybrid Video Coding, Degree Thesis, Institut für Nachrichtentechnik der Rheinisch-Westfälischen Technischen Hchschule Aachen, February 2004.

[38] E. Christophe, *et al*, "Hyperspectral image compression: adapting SPIHT and EZW to anisotopic 3D wavelet coding," submitted to IEEE Transactions on Image processing, pp.1-13, 2006.

[39] L.S. Rodríguez del Río, "Fast piecewise linear predictors for lossless compression of hyperspectral imagery," Thesis for Degree in Master of Science in Electrical Engineering, University of Puerto Rico, Mayaguez Campus, 2003.

[40] -. *Hyperspectral Data Compression,* Edited by Giovanni Motta, Francesco Rizzo and James A. Storer, Chapter 3, Springer, New York, 2006.

[41] B. Arnold, An Investigation into using Singular Value Decomposition as a method of Image Compression, University of Canterbury Department of Mathematics and Statistics, September 2000.

[42] S. Tjoa, *et al*, "Transform coder classification for digital image forensics".

[43] M. Mastriani, and A. Giraldez, "*Smoothing of coefficients in wavelet domain for speckle reduction in Synthetic Aperture Radar images*," ICGST International Journal on Graphics, Vision and Image Processing (GVIP), Volume 6, pp. 1-8, 2005.

[44] M. Mastriani, and A. Giraldez, "*Despeckling of SAR images in wavelet domain*," GIS Development Magazine, Sept. 2005, Vol. 9, Issue 9, pp.38-40.

[45] M. Mastriani, and A. Giraldez, "*Microarrays denoising via smoothing of coefficients in wavelet domain*," WSEAS Transactions on Biology and Biomedicine, 2005.

[46] M. Mastriani, and A. Giraldez, "*Fuzzy thresholding in wavelet domain for speckle reduction in Synthetic Aperture Radar images*," ICGST International on Journal of Artificial Intelligence and Machine Learning, Volume 5, 2005.

[47] M. Mastriani, "*Denoising based on wavelets and deblurring via self- organizing map for Synthetic Aperture Radar images*," ICGST International on Journal of Artificial Intelligence and Machine Learning, Volume 5, 2005.

[48] M. Mastriani, "*Systholic Boolean Orthonormalizer Network in Wavelet Domain for Microarray Denoising*," ICGST International Journal on Bioinformatics and Medical Engineering, Volume 5, 2005.

[49] M. Mastriani, "*Denoising based on wavelets and deblurring via self-organizing map for Synthetic Aperture Radar images*," International Journal of Signal Processing, Volume 2, Number 4, pp.226-235, 2005.

[50] M. Mastriani, "*Systholic Boolean Orthonormalizer Network in Wavelet Domain for Microarray Denoising*," International Journal of Signal Processing, Volume 2, Number 4, pp.273-284, 2005.

[51] M. Mastriani, and A. Giraldez, "*Microarrays denoising via smoothing of coefficients in wavelet domain*," International Journal of Biomedical Sciences, Volume 1, Number 1, pp.7-14, 2006.

[52] M. Mastriani, and A. Giraldez, "*Kalman' Shrinkage for Wavelet-Based Despeckling of SAR Images*," International Journal of Intelligent Technology, Volume 1, Number 3, pp.190-196, 2006.

[53] M. Mastriani, and A. Giraldez, "*Neural Shrinkage for Wavelet-Based SAR Despeckling*," International Journal of Intelligent Technology, Volume 1, Number 3, pp.211-222, 2006.

[54] M. Mastriani, "*Fuzzy Thresholding in Wavelet Domain for Speckle Reduction in Synthetic Aperture Radar Images*," International Journal of Intelligent Technology, Volume 1, Number 3, pp.252-265, 2006.

[55] M. Mastriani, "*New Wavelet-Based Superresolution Algorithm for Speckle Reduction in SAR Images* ," International Journal of Computer Science, Volume 1, Number 4, pp.291-298, 2006.

[56] M. Mastriani, "Denoising and compression in wavelet domain via projection onto approximation coefficients," International Journal of Signal Processing, Volume 5, Number 1, pp.20-30, 2008.

[57] H. Samet, The Design and Analysis of Spatial Data Structures, Addison-Wesley, 1990.

[58] H. Samet, Applications of Spatial Data Structures – Computer Graphics, Image Processing and GIS, Addison-Wesley, 1990.

[59] A.K. Jain, *Fundamentals of Digital Image Processing,* Englewood Cliffs, New Jersey, 1989.

[60] MATLAB® R2008a (Mathworks, Natick, MA).

[61] A. Sharma, and K.K. Paliwal, "Fast principal component analysis using fixed-point algorithm", Pattern Recognition Letters, Vol.28, pp.1151-1155, 2007.

[62] E. Oja, "A Simplified Neuron Model as a Principal Component Analyzer", Journal of Mathematical Biology, Vol.15, pp.267-273, 1982.

[63] T.D. Sanger, "Optimal Unsupervised Learning in a Single-Layer Linear Feedforward Neural Networks," Vol.2, pp.459-473, 1989.